\documentclass[12pt]{article}
\usepackage{lingmacros}
\usepackage[english]{babel}
\usepackage{graphicx}

\large \baselineskip = 20 true pt

\begin{document}
\begin{center}
{\Large \bf The algorithm of the impulse noise filtration in images based on an algorithm 
of community detection in graphs} \vspace{0.5cm}
\end{center}

\begin{center}
S.V. Belim, S.B. Larionov\\
Dostoevsky Omsk State University, Omsk, Russia

 \vspace{0.5cm}
\end{center}

\begin{center}
{\bf Abstract}
\end{center}
This article suggests an algorithm of impulse noise filtration, based on the community 
detection in graphs. The image is representing as non-oriented weighted graph. Each pixel 
of an image is corresponding to a vertex of the graph. Community detection algorithm 
is running on the given graph. Assumed that communities that contain only one pixel 
are corresponding to noised pixels of an image. Suggested method was tested with help of computer 
experiment. This experiment was conducted on grayscale, and on colored images, on artificial 
images and on photos. It is shown that the suggested method is better than median filter by 20\%
regardless of noise percent. Higher efficiency is justified by the fact that most of filters are
changing all of image pixels, but suggested method is finding and restoring only noised pixels. 
The dependence of the effectiveness of the proposed method on the percentage of noise 
in the image is shown.

\section{Introduction}

Impulse noise in the graphical images is looks like random change of  color of some random 
pixels, that called damaged pixels \cite{b1,b2}. The presense of noise in the image affects 
not only on the visual perception, but also affects the results of image segmentation algorithms,
contour highlighting, pattern recognition, etc.

There are a lot of smoothing filters are existing for reduction of noise. Most often used are
Wiener filter and median filter \cite{b3}. One need to point out a non-local filtration methods
\cite{b4,b5,b6}, that demonstrate a better impulse noise reduction. Nevertheless, smoothing 
filters has a significant disadvantage: they changing the whole image and not only the damaged
pixels. This changes are lead to smoothing of contours of the image and make it difficult to
additional processing.

To minimize the impact of filter on the non damaged part of the image, one can use an approach 
based on searching pixels that has changed by the impulse noise. Objective of searching 
the damaged pixels is a very complex problem and solving as a rule by using Data Mining on the 
image. In this approach the algorithm of the impulse noise reduction is consists of two steps. 
The first step is a finding of damaged pixels. And the second is choosing color for each damaged pixel.

Algorithms of damaged pixel searching can be divided into 2 groups. The first group is intended 
to search a Solt \& Pepper Noise (SPN). The feature of this type of noise is that damaged pixels 
can have only maximum or minimum color of the palette. So, the SPN algorithms are based on this 
information \cite{b7,b8,b9}. However, either this type of algorithms is not guarantee that 100\% 
of damaged pixels will be found. The second group is oriented on the impulse noise with random
values. For this group, algorithms based on scheme SD-ROM are most widely used \cite{b10,b11}. 
The main idea of the SD-ROM scheme is that pixels are analyzing within a sliding window with size
of $3\times 3$ and a decision is made about the corruption of the central pixel. The decision
algorithm in SD-ROM is based on the threshold circuit. A decision algorithm, based on the 
hierarchy analysis method has suggested in the paper \cite{b12} wherein the sliding window 
is still using. In the paper \cite{b13} for damaged pixels searching the method of associative
rules has used. In the paper \cite{b14} for damaged pixels searching, an image segmentation
algorithm is using.

Algorithms of impulse noise filtration with a known list of damaged pixels are reduced to filling
out a table with omissions.  The easiest way is to choose color based on colors of the nearest
neighbors. This method is produce good results on the areas with uniform fill, but not acceptable
in case of sharp color transitions, since it leads to blurring of the boundaries. Much more
acceptable approach based on the linear manifolds \cite{b15}. Additionally, the neural networks 
can be used for impulse noise filtration.

As shown by the analysis of previous works, methods based on the analysis of the entire image
provide better impulse noise reduction then local methods. The purpose of this article is to
implement and test an algorithm for impulse noise filtration based on the method of community
detection in graphs, which has proved itself in the task of image segmentation \cite{b16}.

\section{Formulation of the problem and filtration algorithm}

It's assumed that the input of the algorithm is an image damaged by impulse noise with size 
of $NM$ pixels. There is a pair of integers $(x,y)$ is used to determine a location of pixel 
on the image, that represents coordinates of this pixel. The number $x$ takes integer values 
on the interval $[0, N – 1]$, $y$ takes integer values on the interval $[0, M – 1]$. 
In a case of colored image, a pixel with coordinates $(x, y)$ is characterized by 3 color
components: $r(x, y)$ -- intensity of a red color, $g(x, y)$ -- intensity of a green color, 
$b(x, y)$ -- intensity of a blue color.

Let's correspond the weighted, unoriented graph $G$ to the image. Each pixel of the image is
corresponding to a vertex in the graph $G$. Each vertex has an edges only to all nearest 
neighbors. The weight of an edge is calculated based on the color components of connected 
vertexes. For two neighbor vertexes $v_i = (x_i, y_i)$ and $v_j = (x_j, y_j)$, the weight 
of the edge will be equal to:
\[
d(v_i,v_j)=exp\left(-\frac{1}{h}\sqrt{(r_i-r_j)^2+(g_i-g_j)^2+(b_i-b_j)^2} \right),
\]
Here, $r_i = r(x_i, y_j)$, $g_i = g(x_i, y_i)$, $b_i = b(x_i, y_i)$. Parameter $h$ is used 
to change the difference between neighbor pixels, corresponding to moving into another segment 
of the image. This parameter is setting by user and used for the whole image. As it shown 
in paper \cite{b14,b18} this kind of weight function allows to accurately distinguish the 
color change corresponding to the boundaries of the areas in the image. Let's split the graph 
on insets with vertexes that connected much more than others. Such kind of insets are called
communities. A quantitative estimation of the split can be obtained by Newman's modularity 
function \cite{b20, b21}. Greater value of the modularity function means more qualitatively 
the partition is performed. Assumed that damaged pixels are pixels that decrease value of the
modularity function during connection to other communities. This way one can select communities
that contain only one vertex, i.e. vertexes that different from each other a lot. Let's describe
this procedure more formal.

Let's define a matrix of weights E for graph $G$. The values of diagonal elements $E_{ij}$ 
are equal to weight of vertexes. At the first step of the algorithm assume that the weight 
of all vertexes are equal to 0. Other elements of the matrix $E_{ij}$ ($i\neq j$) are equal 
to a weight of a corresponding edge. One should point out that matrix $E_{ij}$ will contain 
a lot of non-zero elements because of only nearest neighbor vertexes are connected with edges 
in graph $G$. The matrix E will be symmetric with respect to the main diagonal because 
of graph $G$ is non-oriented. Let's move to the reduced view of the matrix of weights 
$e = E/m$, where $m=\sum_{i,j=1}^{MN}E_{ij}$. The element $e_{ij}$ is equal to a part 
of edge weight in the whole graph weight. Further one can assume that matrix of weight 
will have a reduced view. It's easy to see that $\sum_{i,j=1}^{MN}e_{ij}=1$.

Modularity is defined as \cite{b21,b22}:
\[
Q(G)=\sum_{i=1}^{K}e_{ii}-\sum_{i=1}^{K}a_ib_i,
\]
where $K$ is a count of vertex in graph, $a_i$ -- reduced outbound power of the vertex $v_i$
($a_i=\sum_{j=1,j\neq i}^{K}e_{ij}$), $b_i$ -- reduced inbound power of the vertex $v_i$
($b_i=\sum_{j=1,j\neq i}^{K}e_{ji}$). Given graph is non-oriented, so outbound and inbound 
power of all vertexes are equal ($a_i = b_i$; $i = 1, ..., K$). Modularity function will have 
a more simpler representation:
\[
Q(G)=\sum_{i=1}^{K}e_{ii}-\sum_{i=1}^{K}a_i^2.
\]

Let's use a screeding procedure to find communities in the graph. Assumed that screed is 
a transformation that replace some inset $H$ of graph $G$ to a some vertex $v_H$. 
If some of vertexes of inset $H$ has been connected by edge with vertex $v$ from 
inset $G\setminus H$, then the vertex $v_H$ will be connected by edge with same weight with 
vertex $v$. The weight of the new edge will be equal to a sum of weights of vertexes and edges that was included to the inset $H$. Let's name a new graph as $G_H$. Let's assume that inset $H$ 
is a community if $Q(G_h) > Q(G)$. Let's point it out that during creation of screed, the count of graph vertexes $(K)$ is decreasing. The main task is to find vertexes that not included to any 
big community. To find such vertexes let's use the following algorithm:

1. Consecutive ñircumvention of all pixels of the image.

2. For each pixel $v$ let's consider nearest neighbor pixels $v^{(i)}$ ($i = 1, ..., 8$). 
Consider insets that contains two vertexes $v$ and $v^{(i)}$ ($i = 1, ..., 8$) 
and try to connect then into community. For each try let's calculate a delta of modularity 
function $\Delta Q_i$ ($i = 1, ..., 8$).

3. If delta of modularity function is negative ($\Delta Q_i<0$, $i = 1, ..., 8$), assumed 
that corresponding pixel is damaged.

The delta of modularity function can be computed fast using current characteristics of the graph
that correspond to an image. For connect to vertexes let's use the following representation 
of the delta of modularity function:
\[
\Delta Q=2(e_{ij}-a_ia_j).
\]

It's obviously that suggested algorithm has linear complexity depending on pixel count.

After finding of damaged pixels one need to choose a color for all of found pixels depending 
on neighbor pixels analysis. Assume that minimal value of the color component of the
neighbor pixels is $m_1$, and maximum value is $m_2$. Let's carry out a sequential search 
all of the values of damaged pixel color from $m_1$ to $m_2$. For each value let's calculate 
delta of the modularity function $\Delta Q$. As a result color let's use the one that takes 
a maximum value of the modularity function delta.

\section{Computer experiment}

Computer experiment has carried out on artificial images of geometric objects and on color photos. 
The value of the impulse noise has been characterized by p, which shows a percent of damaged pixels 
related to a whole pixel count. Impulse noise was generating with help of linear congruent generator 
of pseudo-random numbers. This generator was used both to generate colors and coordinates. During 
computer experiment the percent of damaged pixels has changed from 10\% to 70\%. Image filtration has carried out using suggested method and additionally using a well-known median filter.

To compare the proximity of images there is a Minkovskiy metric \cite{b22, b23} has been used
according to which the proximity between images $A$ and $C$ is calculating using the following
formula:
\[
d(A,C)=\max_{n,m} \sum_{k=1}^{N}\frac{1}{N}\left|A_{nm}^{(K)}-C_{nm}^{(K)} \right|,
\]
where $A_{nm}$ and $Ñ_{nm}$ -- values of $A$ and $Ñ$ image pixel colors, $N$ -- count of pixels.

Relative image enhancement has been calculated based on distance $d(orig, r)$ from reconstructed
image $r$ to original image $orig$ and distance $d(orig, pf)$ from damaged image $pf$ to original image $orig$:
\[
\delta=\frac{d(orig, pf)-d(orig, r)}{d(orig, pf)}\cdot 100\%.
\]

Experiment for a rectangular area with uniform filling showed that the proposed filter allows 
to significantly improve the image. The dependence of the relative improvement on the percentage 
of corrupted pixels for the proposed filter and the median filter is shown in Fig. 1.

\begin{figure}[ht]
\centering
\includegraphics[width=0.6\textwidth]{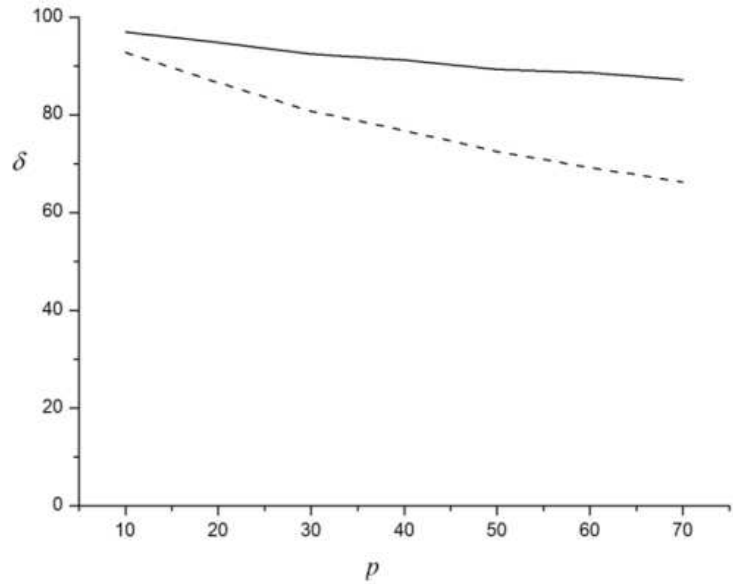}
\caption{Dependence of the relative image improvement on the percentage of noise for the proposed 
filter (solid line) and the median filter (dashed line).}
\label{fig1}
\end{figure}

The results of applying the proposed filter and the median filter to improve the artificial image
with the presence of a solid and gradient fill are shown in Fig. 2.

\begin{figure}[ht]
\centering
\includegraphics[width=0.6\textwidth]{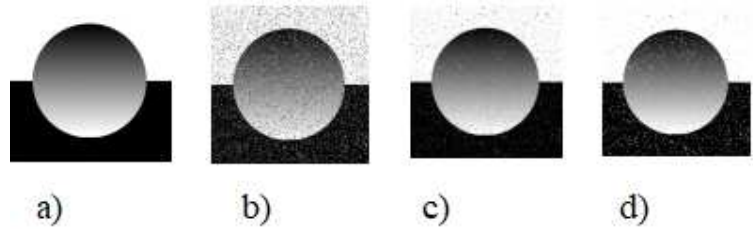}
\caption{The results of applying the filter to an artificial image with a noise level of $p = 20\%$: 
a) the original image, b) a noised image, c) the image reconstructed by the proposed filter, 
d) the image reconstructed by the median filter.}
\label{fig2}
\end{figure}

As can be clearly seen from Fig. 2, the results of the proposed filter are more advantageous for
dark areas of the image, whereas the median filter gives the best visual result for the light 
part of the image. This effect is a related to that new color is choosing for the damaged pixel.
When using the median filter, the colors of the restored pixels are shifted to a white area. In
this case, the colors of the surrounding pixels also change their color. Numerical comparison of
the results of the work shows a significant advantage of the proposed algorithm in front of the
median filter.

The dependence of the relative improvement on the percentage of noised artificial image is shown 
in Figure 3.

\begin{figure}[ht]
\centering
\includegraphics[width=0.6\textwidth]{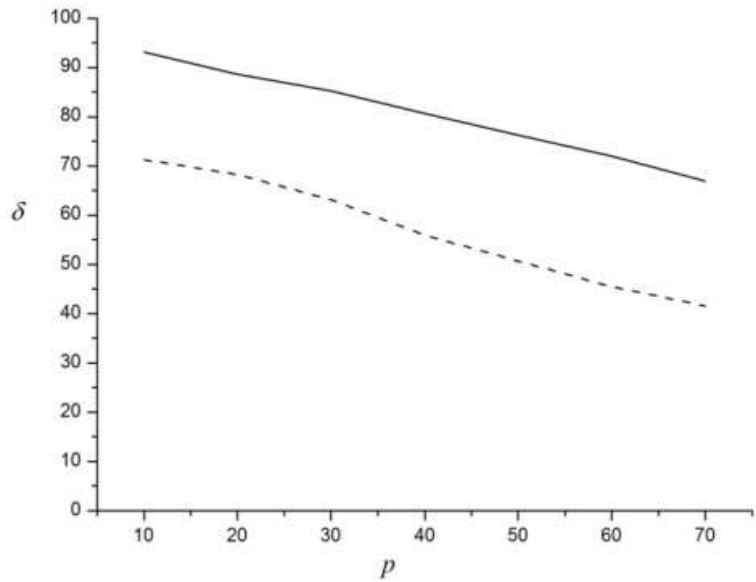}
\caption{Dependence of the relative improvement on the percentage of noised artificial image 
for the proposed filter (solid line) and the median filter (dashed line).}
\label{fig3}
\end{figure}

Additionally the proposed filter allows to get much better results for photographic images. 
The results for the well-known  image "Lena" with a $p = 20\%$ are shown in Fig. 4.

\begin{figure}[ht]
\centering
\includegraphics[width=0.8\textwidth]{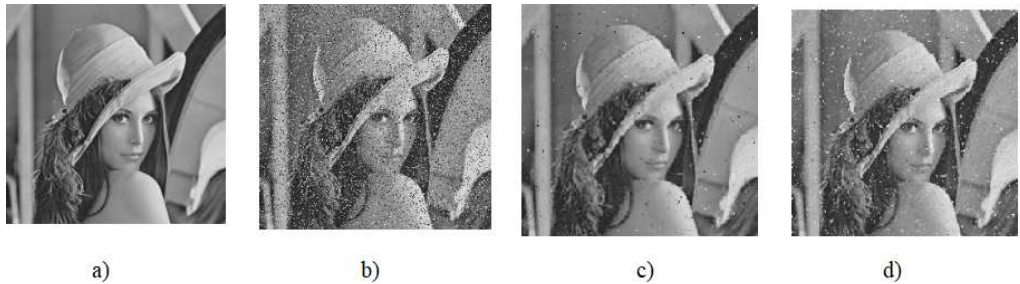}
\caption{The results of application of the filter to the "Lena" image with a noise level of 
$p = 20\%$: a) the original image, b) noised image, c) the image reconstructed by the 
proposed filter, d) the image reconstructed by the median filter.}
\label{fig4}
\end{figure}

It is well known that the "Lena" image is characterized by a large number of small details 
that create difficulties for all filters. As can be seen from Fig. 4, the proposed filter 
gives significantly better results even in a visual comparison. The dependence of the relative
improvement on the percentage of noise is shown in Fig. 5.

\begin{figure}[ht]
\centering
\includegraphics[width=0.6\textwidth]{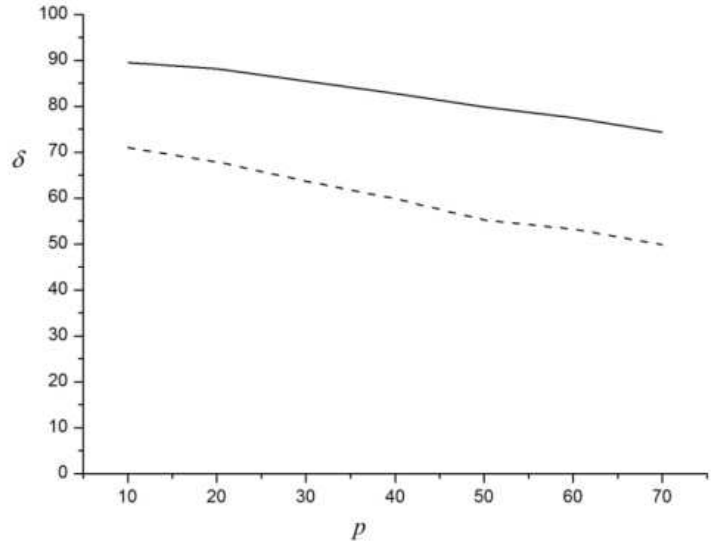}
\caption{Dependence of the relative improvement on the percentage of noise of the image "Lena" 
using the proposed filter (solid line) and the median filter (dashed line).}
\label{fig5}
\end{figure}

\section{Conclusion}

Thus, the proposed filter has good characteristics with linear labor input. As can be seen from 
the graphs presented in Figures 1, 3 and 5, the efficiency of this filter is approximately 20\%
higher than the median for any percentage of damaged pixels. This noticeable advantage is due to
the fact that conventional filters change all pixels of the image. Correcting the damaged pixels
brings the image closer to the original, but changing the undamaged pixels increases the distance
to the original. This property is inherent not only to the median filter, but also to all
traditional filters.

The filter proposed in this article acts selectively and changes only those pixels that differ
significantly from those around them. With a high probability, such pixels will be damaged by
impulse noise. The choice of a new color on the basis of attaching to one of the neighboring
communities of pixels allows to form communities of pixels close in characteristics. It should 
also be noted the high speed of the filter, due to the linear complexity of the algorithm
underlying it. The processing time of one image within the error is the same as the median filter.


\end{document}